## ARTICLE INFORMATION

**Authors**

Thanadol Singkhornart[a], Olarik Surinta[a,*]

**Affiliations**

[a] Multi-agent Intelligent Simulation Laboratory (MISL) Research Unit, Department of Information Technology, Faculty of Informatics, Mahasarakham University, Mahasarakham 44150, Thailand

**Corresponding author's email address**

Email address: olarik.s@msu.ac.th (O. Surinta)

**Abstract**

The Multi-language Video Subtitle Dataset is a comprehensive collection designed to support research in text recognition across multiple languages. This dataset includes 4,224 subtitle images extracted from 24 videos sourced from online platforms. It features a wide variety of characters, including Thai consonants, vowels, tone marks, punctuation marks, numerals, Roman characters, and Arabic numerals. With 157 unique characters, the dataset provides a resource for addressing challenges in text recognition within complex backgrounds. It addresses the growing need for high-quality, multilingual text recognition data, particularly as videos with embedded subtitles become increasingly dominant on platforms like YouTube and Facebook. The variability in text length, font, and placement within these images adds complexity, offering a valuable resource for developing and evaluating deep learning models. The dataset facilitates accurate text transcription from video content while providing a foundation for improving computational efficiency in text recognition systems. As a result, it holds significant potential to drive advancements in research and innovation across various computer science disciplines, including artificial intelligence, deep learning, computer vision, and pattern recognition.

# SPECIFICATIONS TABLE

| Subject | *Computer Science* |
|---|---|
| Specific subject area | The multi-language video subtitle image dataset consists of images containing text in various languages, specifically designed to facilitate text recognition within images. This dataset is relevant to several disciplines, including artificial intelligence, deep learning, computer science applications, computer vision, and pattern recognition. |
| Type of data | Image (JPG format) |
| Data collection | The multi-language video subtitle images were sourced from online platforms containing content in various languages, including Thai characters, English characters, Thai numerals, Arabic numerals, and special symbols. The dataset was created by extracting frames from videos, followed by manually annotating the subtitle locations within each frame. As a result, a single frame may contain multiple subtitle regions. |
| Data source location | Online platforms: YouTube and Facebook |
| Data accessibility | Repository name: Mendeley Data<br>Data identification number: 10.17632/gj8d88h2g3.2<br>Direct URL to data: https://data.mendeley.com/datasets/gj8d88h2g3/2 |
| Related research article | T. Singkhornart, O. Surinta, Multi-language video subtitle recognition with convolutional neural network and long short-term memory networks, ICIC Express Letters 16 (2022) 647–655. https://doi.org/ 10.24507/icicel.16.06.647 |

# VALUE OF THE DATA

- The Multi-language Video Subtitle Dataset provides a substantial sample size of 4,224 subtitle images, focusing on two primary languages, Thai and English, as well as two numeral systems (Thai and Arabic) and special characters, encompassing a total of 157 distinct characters. Labels corresponding to the subtitle content are embedded within the filenames, enabling efficient referencing and organization. A single sample image may contain a mix of Thai, English, and numerals, presenting challenges in recognizing multiple languages simultaneously.
- The text within the images varies significantly, with some appearing against complex backgrounds. In this dataset, the shortest text length is one character, while the longest extends to approximately 80 characters. Additionally, around 80 samples contain texts ranging from 10 to 40 characters.
- Researchers in computer science and related fields can utilize this dataset to advance research and enhance performance in terms of accuracy and computational efficiency. Moreover, various deep learning methodologies can be applied to further explore this dataset and push the boundaries of text recognition research.

## BACKGROUND

Massive numbers of videos are uploaded to online platforms like YouTube and Facebook, making them mainstream channels, particularly among teenagers. Many popular TV channels worldwide have embraced this trend, offering high-quality content through these platforms. Numerous videos also provide subtitles, which are necessary for individuals with hearing impairments, enabling them to comprehend the video content. Subtitles further assist audiences in learning the spelling of names, brands, acronyms, and abbreviations and understanding various accents. Additionally, subtitles bridge the accessibility gap between content and audiences. Further supporting non-native speakers, subtitles serve as essential tools for individuals with hearing impairments, offering a textual representation of spoken dialogue that transcends language barriers and audio limitations.

In this dataset, the research team focused on capturing subtitles from online platforms like YouTube and Facebook that feature multiple languages, including Thai and English. The text within these images consists of a mix of Thai, English, and numerals, often displayed against complex backgrounds, making recognition particularly challenging. It emphasizes the requirement for high-quality datasets that provide comprehensive data for developing deep learning models and enhancing text recognition systems. Advanced recognition systems rely on robust models capable of accurately transcribing text from videos, even when encountered with font, size, and language variations.

## DATA DESCRIPTION

The Multi-language Video Subtitle Dataset is a comprehensive collection of images containing text in multiple languages, referred to as subtitle images. These images were extracted from 24 videos sourced from online platforms. The details of the dataset are presented as follows.

**Character Collection**

The subtitle images were extracted from 24 videos containing a diverse range of characters across multiple languages. These include Thai consonants, vowels, tone marks, punctuation marks, and numerals, as well as Roman characters and Arabic numerals. The Multi-language Video Subtitle Dataset comprises 157 unique characters, as detailed in Table 1.

**Table 1.** Character collection in the Multi-language Video Subtitle Dataset.

| Character Types | | Characters |
|---|---|---|
| Thai | Consonant | ก ข ฃ ค ฅ ฆ ง จ ฉ ช ซ ฌ ญ ฎ ฏ ฐ ฑ ฒ ณ ด ต ถ ท ธ น บ ป ผ ฝ พ ฟ ภ ม ย ร ล ว ศ ษ ส ห ฬ อ ฮ |
| | Vowel | ะ ั า ำ ิ ี ึ ื ุ ู ำ ํ ใ ไ โ ๐ เ แ ฤ ฦ |
| | Tone | ่ ้ ๊ ๋ |
| | Punctuation mark | ๆ ฯ ์ |
| | Numeral | ๑ ๒ ๓ ๔ ๕ ๖ ๗ ๘ ๙ ๐ |
| Roman | Character | A B C D E F G H I J K L M N O P Q R S T U V W X Y Z a b c d e f g h i j k l m n o p q r s t u v w x y z |
| Arabic | Numeral | 1 2 3 4 5 6 7 8 9 0 |
| Special character | | . , ! ( ) - $ ฿ & : ? * " " ' ' (space) |

**Labelling**

The research team split the 24 videos into frames, resulting in 4,224 images with a resolution of 1,280×720 pixels per frame. Out of the total, 2,700 images containing subtitles were selected and annotated. However, the dataset does not include annotation files (XML); only the image files (JPG) are freely accessible via the Mendeley Data repository. Examples of subtitle images and their labels are illustrated in Table 2.

**Table 2.** Examples of images with subtitles labels.

| Images with subtitles | Labels |
|---|---|
| 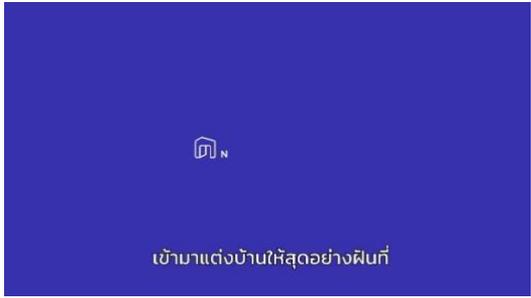 | เข้ามาแต่งบ้านให้สุดอย่างฝันที่ |

| Images with subtitles | Labels |
|---|---|
| 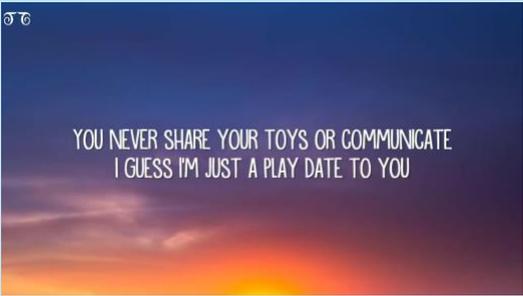 | YOU NEVER SHARE YOUR TOYS OR COMMUNICATE<br><br>I GUESS I'M JUST A PLAY DATE TO YOU |
| 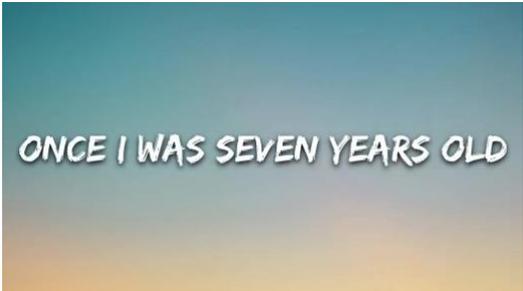 | ONCE I WAS SEVEN YEARS OLD |

**File format and conversion**

After the annotation and labelling process, a Python program was developed to extract the subtitle images based on the coordinates provided in the XML files. The extracted subtitle images are stored in format with RGB channels, ensuring compatibility with widely used image processing libraries and tools.

Furthermore, the filenames of each subtitle image follow a consistent and standardized naming convention, embedding both a unique identifier and a label corresponding to the subtitle content. Examples of subtitle images and their labels are presented in Table 3.

**Table 3.** Examples of subtitle images and labels.

| Subtitle Images | Filenames |
|---|---|
| 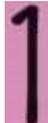 | 3798_1.jpg |

| Subtitle Images | Filenames |
|---|---|
| 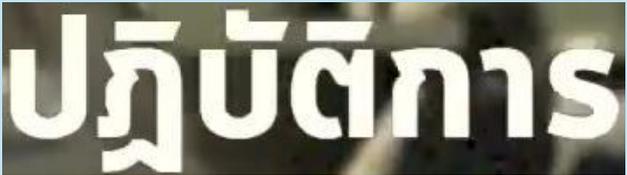 | 3148_ปฏิบัติการ.jpg |
| 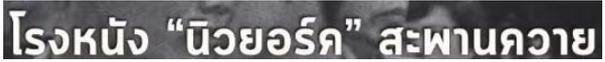 | 1001_โรงหนัง นิวยอร์ค สะพานควาย.jpg |
| 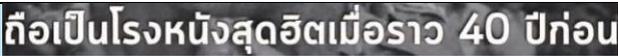 | 1002_ถือเป็นโรงหนังสุดฮิตเมื่อราว 40 ปีก่อน.jpg |
| 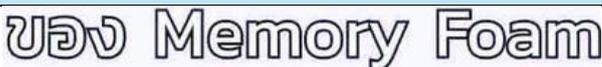 | 105_ของ Memory Foam.jpg |
| 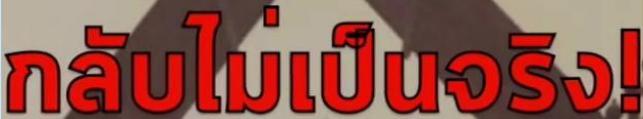 | 1149_กลับไม่เป็นจริง!.jpg |
| 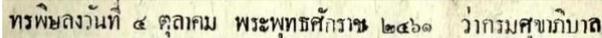 | 2190_ทรพิษลงวันที่ ๔ ตุลาคม พระพุทธศักราช ๒๔๖๑ ว่าด้วยกรมศุขาภิบาล.jpg |
| 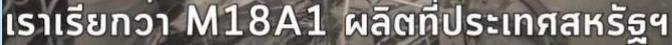 | 3374_เราเรียกว่า M18A1 ผลิตที่ประเทศสหรัฐฯ.jpg |
| 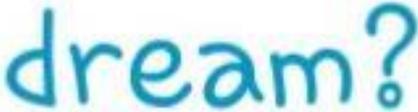 | 4205_dream_.jpg |
| 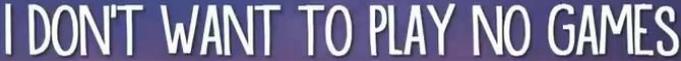 | 4041_I DON_T WANT TO PLAY NO GAMES.jpg |

46

**Distribution of characters and subtitle images**

After generating the subtitle images and labels, we calculated basic statistics for the characters, as presented in Fig. 1 and Fig. 2.

As depicted in Fig. 1, the distribution of characters shows the number of subtitle images relative to the number of characters appearing in each image. It reveals that more than 80 subtitle images contain between 10 and 40 characters, which represents the standard length for subtitles. However, a small number of images contain fewer than 10 characters or more than 40 characters, underscoring the importance of balanced training to ensure models can effectively handle both shorter and longer subtitles.

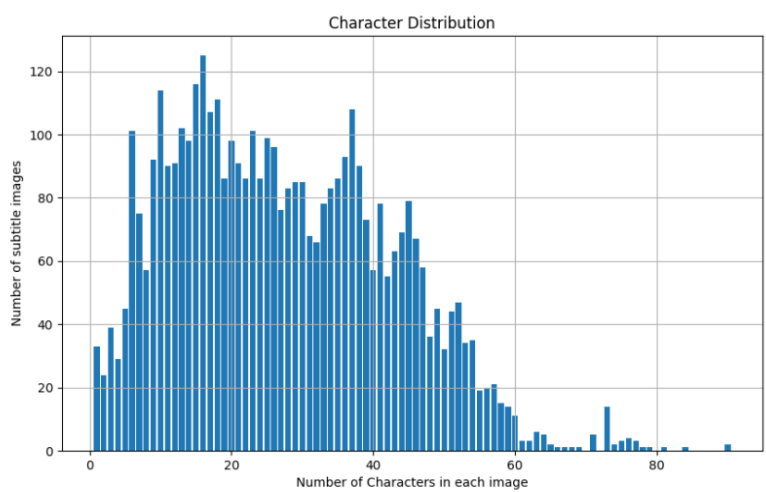

**Fig. 1.** illustrates the number of subtitle images (y-axis) relative to the number of characters appearing in each image (x-axis).

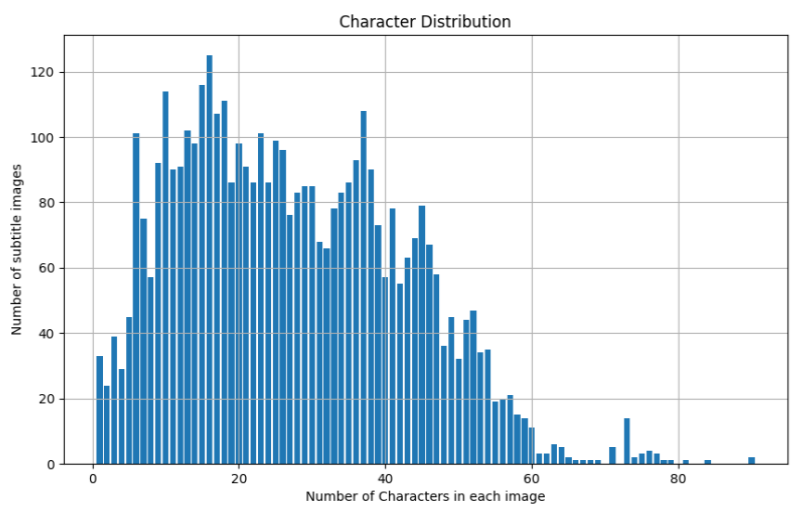

**Fig. 2.** illustrates the number of pixels in both the width (blue) and height (red) of the subtitle images (x-axis) relative to the number of subtitle images (y-axis). It is recommended to view this graph in color for better clarity.

Additionally, Fig. 2 presents the frequency distribution of image dimensions. The red line, representing image height, shows a prominent peak at lower pixel values, indicating that most images have a shorter height. The blue line, representing image width, displays a broader distribution, with widths extending up to 1,200 pixels. Moreover, Fig. 3 highlights the maximum width and height of the subtitle images in the Multi-language Video Subtitle Dataset.

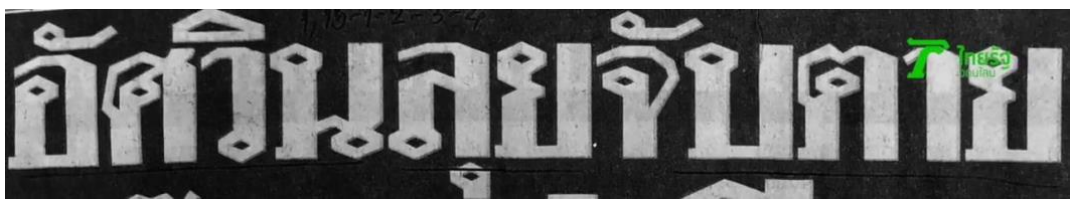

(a) 1,279×229 pixels

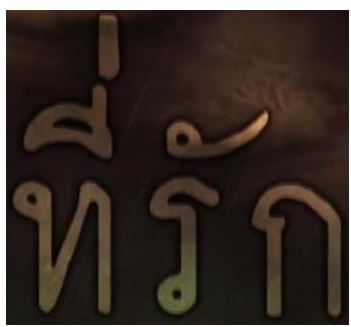

(b) 675×624 pixels

**Fig. 3.** illustrates subtitle images with width×height dimensions, showing (a) a maximum width of 1,279 pixels and (b) a maximum height of 624 pixels collected in the dataset.

## EXPERIMENTAL DESIGN, MATERIALS AND METHODS

**Materials**

The subtitle images were collected from YouTube and Facebook between October 2020 and January 2021 by selecting videos that featured embedded subtitles. In the initial phase of data collection, the research team focused solely on subtitles appearing at the bottom of the video. However, it was later observed that many videos, particularly those from news channels, also show graphical text on the screen. Consequently, lyric videos featuring multiple lines of text in the middle of the screen were also included.

The video sources were gathered from the following Facebook pages: ไทยรัฐออนไลน์ (3นาทีคดีดัง) and TEP - Thailand Education Partnership ภาคีเพื่อการศึกษาไทย. The YouTube channels included Bearhug, KLUAYTHAI, KRIT Eighth Floor, Genierock, Taj Tracks, Cakes & Eclairs, 7clouds, DopeLyrics, Equilanora, JEMIN Apollo, San Ko, Tangerine JJY, SNH48 Lyrics, and Lemoring. After downloading the videos, they were split into frames, with a frame captured every 5 seconds, providing a comprehensive snapshot of the video content. An example of images with subtitle labels is shown in Fig. 4.

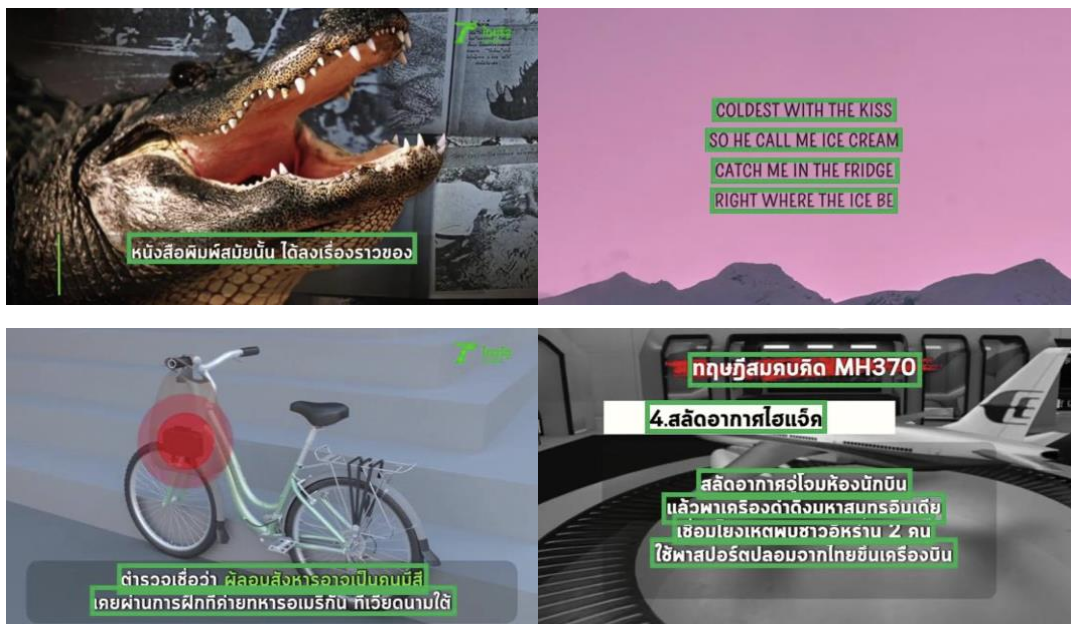

**Fig. 4.** Example images with subtitles. Note that the green rectangle boxes represent the locations of the subtitles.

The LabelImg software was used to annotate the subtitles in the images (see Fig. 5), generating XML files with precise subtitle locations and corresponding labels. A Python program was then developed to extract the subtitle images and labels based on the information provided in the XML files.

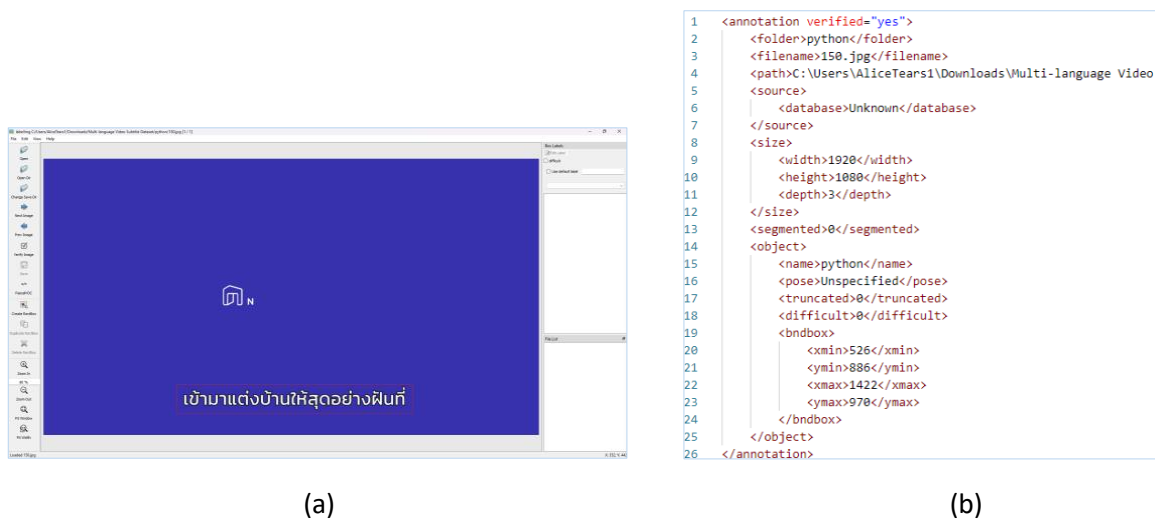

(a) (b)

**Fig. 5.** Examples of (a) the LabelImg software and (b) the XML file, which provides the location of subtitles.

Moreover, the subtitle images include text with various fonts, sizes, and colors. The number of characters in each subtitle image also varies. As a result, some subtitle images contain both Thai and English characters. Example subtitle images and their dimensions are shown in Table 4.

**Table 4.** Example subtitle images and their dimensions.

| Subtitle images and dimensions |
| --- |
| 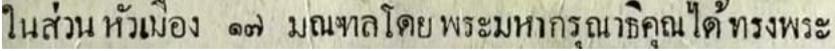<br>623×39 pixels |
| 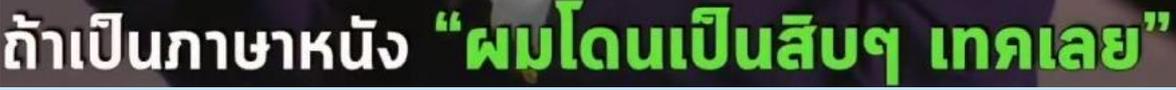<br>878×68 pixels |
| 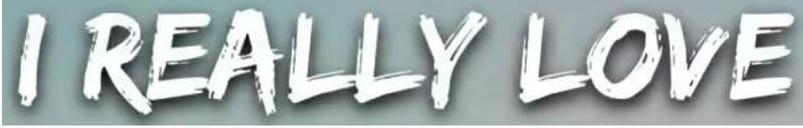<br>598×96 pixels |
| 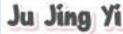<br>92×99 pixels |
| 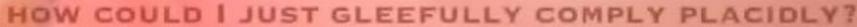<br>624×25 pixels |
| 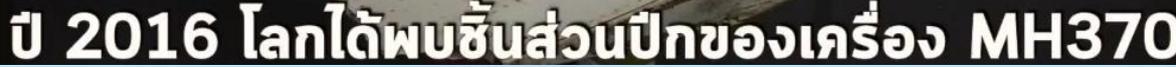<br>997×61 pixels |

**Experimental design**

Deep learning methods, such as convolutional neural networks (CNNs) and long short-term memory (LSTM) networks, have been proposed to evaluate model performance using the Multi-language Video Subtitle Dataset. In 2022, Singkhornart and Surinta [1] introduced a CNN-LSTM architecture where the output utilized connectionist temporal classification (CTC) as the loss function to compute the loss value and decode the predictions. The modified VGG19 was employed as the CNN backbone, resulting in a character error rate (CER) of 9.36%.

Additionally, Gonwirat, Surinta, & Pawara [2] proposed the FusionCNNs-LSTM architecture to extract more robust temporal features from the subtitle images. Their approach integrated the fusion of VGG-s1 and VGG-s2 architectures, where s represents the stride number, using an additive operation. They also compared two decoding algorithms, CTC and word beam search (WBS), for decoding the output. Their experimental results indicated that WBS outperformed the CTC algorithm, with the FusionCNNs-LSTM achieving a CER of 5.29% on the test set.

Both of these methods focus on the fusion of deep learning algorithms, particularly CNN and LSTM architectures, to extract both spatial and temporal features. Additionally, to enhance feature robustness, a multi-layer adaptive spatial-temporal feature fusion network (ASTFF) [3] can be employed to extract spatial-temporal features before passing them to the decoding algorithms.

To further advance text recognition, several methods, such as LISTER [4], CLIPTER [5], and Class-Aware Mask-guided [6] have been developed to address varying text lengths and challenging conditions, such as handwritten, blurred, and noisy text. These approaches not only prioritize accuracy but also emphasize computational efficiency, with a strong focus on minimizing processing costs.

## LIMITATIONS

Not applicable

## ETHICS STATEMENT

The authors have read and follow the ethical requirements and confirming that the current work does not involve human subjects, animal experiments, or any data collected from social media platforms.

## CRediT AUTHOR STATEMENT

**Thanadol Singkhornart:** Conceptualization, Data Curation, Investigation, Methodology, Resources, Validation, Writing – Original Draft; **Olarik Surinta:** Supervision, Conceptualization, Experimental Design, Writing – Review & Editing, Funding Acquisition.

## ACKNOWLEDGEMENTS

This research project was financially supported by Mahasarakham University, Thailand

## DECLARATION OF COMPETING INTERESTS

The authors declare that they have no known competing financial interests or personal relationships that could have appeared to influence the work reported in this paper.